\documentclass[11pt]{article}

\usepackage[final]{acl}

\usepackage{times}
\usepackage{latexsym}
\usepackage{booktabs}
\usepackage{tabularx}
\usepackage{amssymb}

\usepackage[T1]{fontenc}

\usepackage[utf8]{inputenc}

\usepackage{microtype}

\usepackage{inconsolata}

\usepackage{graphicx}

\usepackage{amsmath}
\usepackage{amssymb}
\usepackage{mathtools}
\usepackage{amsthm}
\usepackage{algorithm}
\usepackage{algorithmic}
\usepackage{colortbl}
\usepackage{array}
\usepackage{multirow}
\usepackage{subfig} 
\usepackage{booktabs} 

\title{Routing-Based Continual Learning for Multimodal \\ Large Language Models}

\author{Jay Mohta\thanks{Equal contribution} , \text{ }Kenan Emir Ak\footnotemark[1] , \text{ }Gwang Lee, \text{ } Dimitris Dimitriadis, \text{ }Yan Xu, \text{ }Mingwei Shen\\
  Amazon.com\\
  {\tt \{jaymoht, kenanea, gglee, dbdim, yanxuml, mingweis\}@amazon.com}}

\begin{document}
\maketitle
\begin{abstract}
Multimodal Large Language Models (MLLMs) struggle with continual learning, often suffering from catastrophic forgetting when adapting to sequential tasks. We introduce a routing-based architecture that integrates new capabilities while robustly preserving foundational knowledge. While Multi-Task Learning (MTL) offers a theoretical performance upper bound, it incurs a linearly scaling computational overhead as the number of tasks increases. In contrast, our method maintains fixed data and compute requirements regardless of the task sequence length. Across models ranging from 2B to 8B parameters, we demonstrate that our routing approach performs on par with MTL while retaining the training efficiency of sequential fine-tuning. Beyond merely mitigating forgetting, we observe that token-level routing facilitates cross-modal transfer, leveraging knowledge from one modality to bolster performance in another. Ablation studies confirm the approach's scalability: routing remains robust even with large expert pools and effectively capitalizes on task relatedness. Finally, we show that our method scales favorably, with larger models exhibiting minimal degradation compared to fully specialized fine-tuning.
\end{abstract}

\section{Introduction}

Large Language Models (LLMs) have fundamentally transformed Natural Language Processing (NLP)~\cite{brown2020language}, a success now mirrored by Multimodal Large Language Models (MLLMs) capable of reasoning leveraging visual and textual inputs~\cite{radford2021learning,alayrac2022flamingo,liu2023visual}. However, adapting these general-purpose models to specialized domains introduces a critical dilemma: optimizing performance on target tasks via fine-tuning frequently triggers catastrophic forgetting (CF), severely degrading the model’s previously acquired knowledge~\cite{mccloskey1989catastrophic}.

This challenge is most severe in continual learning settings, where models must learn a sequence of tasks without access to historical training data~\cite{kemker2018measuring}. Given the massive scale of MLLMs, traditional solutions such as experience replay~\cite{rolnick2019experience,rebuffi2017icarl} or parameter regularization~\cite{kirkpatrick2017overcoming} are often computationally prohibitive or ineffective~\cite{wu2024continuallearninglargelanguage}. Because general linguistic and visual knowledge is distributed across billions of parameters, task-specific updates may lead to catastrophic forgetting, where the optimization for new objectives inadvertently overwrites the representations necessary for earlier tasks~\cite{kotha2023understanding}.


\begin{figure}[t]
    \centering
    \includegraphics[width=1.05\columnwidth]{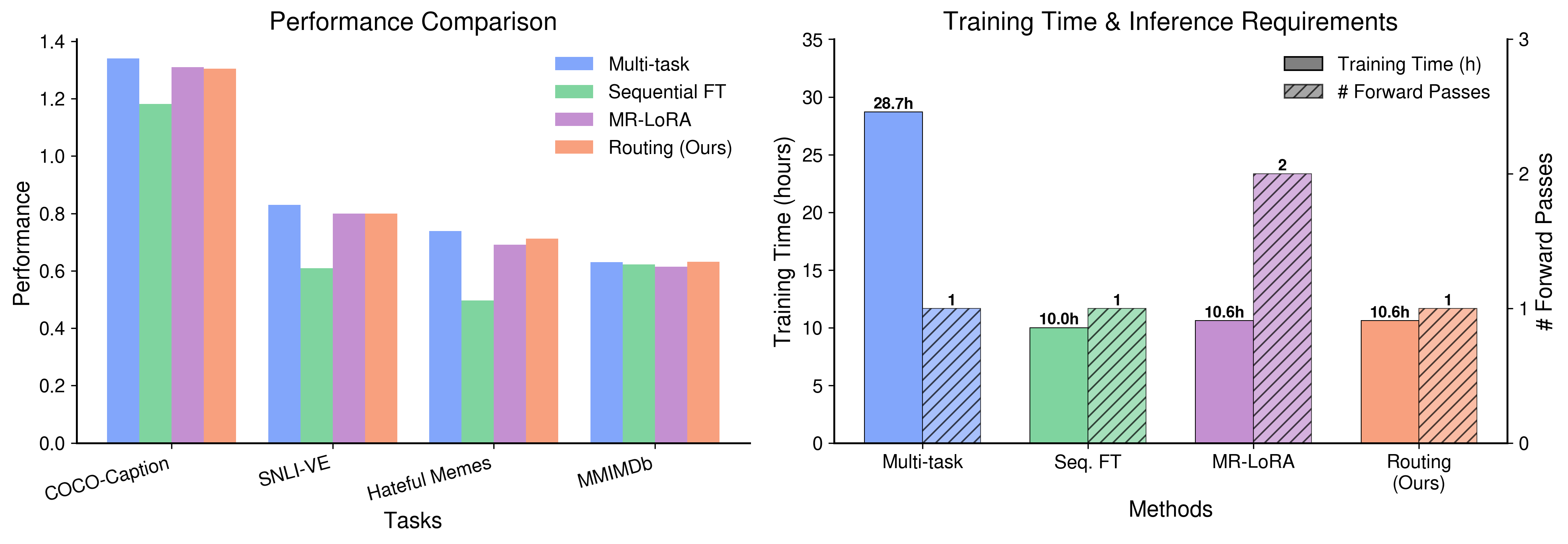}
    \caption{Comparison of continual learning methods. \textbf{Left:} Task performance across four vision-language benchmarks. \textbf{Right:} Training time and inference requirements. Our routing-based approach performs on par with Multi-Task Learning performance (upper bound) while requiring $2.7\times$ less training time and only a single forward pass at inference, unlike MR-LoRA which requires two.}
    \label{fig:comp_multi_task_main_fig}
\end{figure}

In the multimodal context, this interference is particularly damaging as it risks the cross-modal alignment established during pretraining~\cite{zhai2023investigating}. Fine-tuning can decouple visual embeddings from their textual counterparts, breaking the model's reasoning capabilities~\cite{zhu2024beyond}. Furthermore, visual encoders and language decoders often suffer from asymmetric forgetting rates, further destabilizing the model~\cite{ni2023continual}. Consequently, there is a critical need for architectures that allow plasticity for new tasks while enforcing stability for foundational cross-modal alignments.

Figure~\ref{fig:comp_multi_task_main_fig} illustrates the trade-offs between these paradigms. While Sequential Fine-Tuning is data-efficient, it suffers from a sharp decline in performance on prior tasks. Conversely, Multi-Task Learning (MTL) ensures stability by retraining on the cumulative dataset, but incurs a linearly scaling computational cost that is unsustainable for large-scale MLLMs. Furthermore, while modular frameworks like MR-LoRA~\cite{zhao2025mllmclcontinuallearningmultimodal} reduce interference, they often require two forward passes at inference to resolve task-specific outputs. Our approach aims to occupy a middle ground: it achieves the stability of MTL and the efficiency of sequential methods while maintaining only a single-pass inference.

In this work, we introduce a routing-based modular architecture designed to tackle the challenges of continual learning in MLLMs. By leveraging token-level routing~\cite{muqeeth2024learning}, our method isolates task-specific adaptations to prevent the overwriting of shared knowledge. Our contributions are fourfold:
\begin{enumerate}
    \item We demonstrate that routing-based continual learning matches the performance stability of Multi-Task Learning using only the current task's data, eliminating the need for costly retraining.
    \item We uncover that token-level routing enables distinct cross-modal transfer, allowing improvements in unimodal experts (e.g., text-only) to bolster performance in multimodal domains.
    \item We provide extensive evaluations across sequential tasks, demonstrating robustness to expert pool size and the positive utilization of task relatedness.
    \item We identify a positive scaling trend: as model size increases, our method approaches the performance of fully specialized fine-tuning, validating its viability for large-scale deployment.
\end{enumerate}

\section{Related Work}
\subsection{Multimodal Large Language Models}
MLLMs extend the capabilities of LLMs by processing visual inputs alongside text, typically comprising a pretrained vision encoder, a language backbone, and a connector module that projects visual features into the language embedding space~\cite{yin2024survey}. Early architectures, such as Flamingo~\cite{alayrac2022flamingo} and BLIP-2~\cite{li2023blip}, established foundational design principles by bridging frozen vision and language models via modular components.

Subsequent research has prioritized scaling model capacity and refining instruction tuning. LLaVA~\cite{liu2023visual, liu2024improved} demonstrated the efficacy of visual instruction tuning using simple projection layers, while models such as Qwen-VL~\cite{bai2023qwenvl} and InternVL~\cite{chen2024internvl, chen2024far} introduced multi-stage training paradigms incorporating diverse data mixtures and dynamic resolution processing. However, these architectures are typically optimized for static training pipelines under the assumption of i.i.d. data distributions. Consequently, how to effectively incorporate continual adaptation into these frameworks remains unclear; without explicit mechanisms to manage sequential learning, they remain susceptible to catastrophic forgetting.

\subsection{Continual Learning in Large Models}
Catastrophic forgetting remains a persistent challenge in deep learning, exhibiting a relationship with model scale that is both complex and task-dependent~\cite{ramasesh2021effect, kalajdzievski2024scaling}. In MLLMs, this difficulty is intensified by cross-modal interference: sequential adaptation often disrupts the alignment established during pretraining, causing fine-tuning in one modality to degrade representations in the other~\cite{zhai2023investigating}.

Traditional continual learning strategies face severe scalability bottlenecks when applied to billion-parameter models. Replay-based methods~\cite{rolnick2019experience, chaudhry2019tiny} require the retention of historical data, raising privacy and storage concerns, while regularization techniques~\cite{kirkpatrick2017overcoming, zenke2017continual} frequently demand computationally prohibitive importance scoring across the entire network. Although multi-task learning provides a theoretical upper bound for performance, it necessitates simultaneous access to all datasets, rendering it impractical for dynamic data streams~\cite{ruder2017overview}.

To address these limitations, recent Parameter-Efficient Fine-Tuning (PEFT) methods have been adapted for the continual learning setting. Approaches such as Progressive Prompts~\cite{razdaibiedina2023progressive} utilize task-specific tokens, while O-LoRA~\cite{wang2023orthogonal} constrains adaptation to orthogonal subspaces. Similarly, SEMA~\cite{wang2025self} introduces adapter-based sequential training with automatic expansion for vision-based learning. However, the MLLM-CL benchmark~\cite{zhao2025mllmclcontinuallearningmultimodal} indicates that while these approaches mitigate forgetting, they struggle with complex, domain-specific shifts. Unlike these adapter-centric strategies, our method operates at the \textit{token level}, enabling fine-grained routing that better accommodates vision-language tasks.
\subsection{Model Merging and Modular Routing}
Modular architectures offer a promising alternative to rigid end-to-end training. Model merging, for instance, combines the weights of separately fine-tuned models without requiring concurrent data access~\cite{matena2022merging, ilharco2022editing, yadav2023ties}. While computationally efficient, merging is static; it often underperforms when tasks are semantically distant or require conflicting reasoning patterns, leading to interference in the final averaged weights.

Routing-based methods address these challenges by dynamically selecting specialized sub-networks based on input semantics. While frameworks like PHATGOOSE~\cite{muqeeth2024learning} and Arrow~\cite{ostapenko2024towards} utilize modular routing for zero-shot generalization, and Mixture-of-Experts (MoE) architectures~\cite{lin2024moe, riquelme2021scaling} employ it for capacity scaling, their application in continual learning remains underexplored. Specifically, standard MoEs often lack mechanisms to isolate task-specific knowledge or handle the sequential arrival of data. Our approach addresses this by leveraging routing to manage an ever-growing library of Low-Rank Adapters (LoRA). Rather than updating a static set of parameters, we instantiate a new LoRA 'specialist' for each incremental task. The router then learns to steer inputs toward the appropriate module, allowing the MLLM to broaden its capabilities while maintaining strict functional isolation to prevent catastrophic forgetting.

\section{Proposed Method}
\label{sec:proposed_method}
In this paper, we introduce a modular routing framework designed to address the stability-plasticity dilemma in MLLMs~\cite{shi2024continual}. While modular architectures have demonstrated success in zero-shot generalization for text-only LLMs~\cite{muqeeth2024learning, ostapenko2024towards}, adapting this paradigm to multimodal continual learning requires addressing cross-modal interference. Our core insight is the decoupling of \textit{representation learning (skill acquisition)} from \textit{task identification (routing)}. By isolating these processes, we ensure that new capabilities are integrated without eroding the model’s existing cross-modal alignment.

\begin{figure*}[t]
    \centering
    \includegraphics[width=\textwidth]{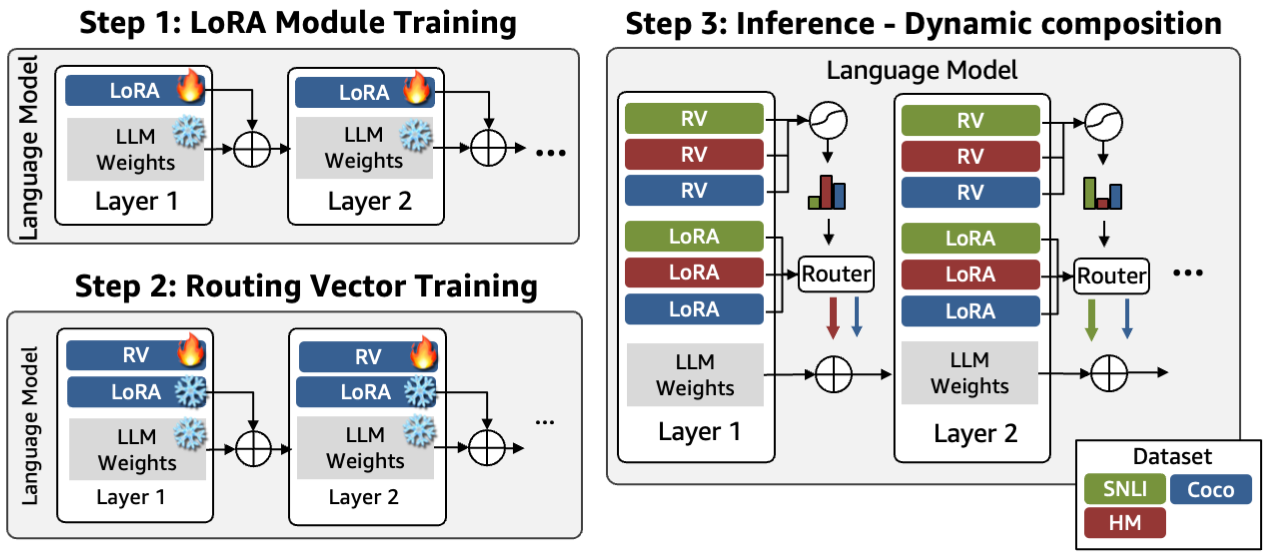} 
    \caption{\textbf{The proposed method.} The process is divided into three steps: (1) Training task-specific LoRA adapters; (2) Training lightweight Routing Vectors (RV) to recognize task-specific activation patterns; (3) Inference using dynamic composition to select the top-$k$ experts per token.}
    \label{fig:method_arch}
\end{figure*}

As illustrated in Figure~\ref{fig:method_arch}, our framework operates in three distinct phases for each sequential task: (1) Expert Specialization via LoRA, (2) Router Vector Training, and (3) Dynamic Inference.

\subsection{Architecture Overview}
We utilize a standard MLLM architecture composed of a vision encoder, a projection layer, and an LLM. To isolate the impact of our routing mechanism and ensure a controlled evaluation of catastrophic forgetting, we keep the vision encoder and projection layer frozen. We implement task-specific LoRA modules within the linear layers of the LLM.

\subsection{Training Strategy}
Our training strategy separates the acquisition of task knowledge from the task selection. This separation is critical for preventing the gradient conflict typically observed in sequential fine-tuning.

\paragraph{Step 1: Expert Specialization (LoRA).}
For each arriving task $T_i$, we specialize the LLM by training a dedicated LoRA adapter. Let $W_0 \in \mathbb{R}^{d \times d_{in}}$ denote the frozen weights of a linear layer in the base LLM. For an input activation $u_t \in \mathbb{R}^{d_{in}}$ at token $t$, the output $h_t$ is computed as:
\begin{equation}
    h_t = W_0 u_t + B_i A_i u_t,
\end{equation}
where $A_i \in \mathbb{R}^{r \times d_{in}}$ and $B_i \in \mathbb{R}^{d \times r}$ are the trainable LoRA parameters with rank $r \ll d$. These parameters are optimized using only the dataset for task $T_i$, ensuring the expert captures task-specific nuances without being regularized by prior tasks.

\paragraph{Step 2: Routing Vector (RV) Training.}
Once the expert for task $T_i$ is trained, we freeze parameters $A_i$ and $B_i$ and add a lightweight routing vector $v_i \in \mathbb{R}^{d_{in}}$. This vector serves as a soft gate, learning to identify the activation subspace associated with task $T_i$.

The routing vector is initialized to zero and optimized using the same next-token prediction objective as Step 1. During this phase, the layer output is governed by:
\begin{equation}
    h_t = W_0 u_t + \sigma(v_i^T u_t) \cdot (B_i A_i u_t),
\end{equation}
where $\sigma(\cdot)$ is the sigmoid function. By optimizing $v_i$ while the expert is active, the router learns a latent ``signature'' of the task within the activation space. Crucially, this step requires no access to data or modules from prior tasks.

\subsection{Inference}
\paragraph{Step 3: Dynamic Inference.}
During inference, we utilize a bank of trained experts and their corresponding routing vectors $\mathcal{V} = \{v_1, \dots, v_N\}$ for $N$ accumulated tasks. As depicted in Step 3 of Figure~\ref{fig:method_arch}, routing decisions are computed dynamically at the token level, allowing the model to switch contexts mid-generation.

For an input activation $u_t$, we compute an affinity score with each routing vector:
\begin{equation}
    s_{t,j} = v_j^T u_t, \quad j \in \{1, \dots, N\}.
\end{equation}
We then identify the set of indices $\mathcal{E}_t$ corresponding to the top-$k$ affinity scores. To effectively mix the selected experts, we normalize these scores using a softmax function with a scaling temperature $\tau$:
\begin{equation}
    w_{t,j} = \frac{\exp(s_{t,j} / \tau)}{\sum_{m \in \mathcal{E}_t} \exp(s_{t,m} / \tau)}, \quad \forall j \in \mathcal{E}_t.
\end{equation}
The final output is the superposition of the base model and the weighted experts:
\begin{equation}
    h_t = W_0 u_t + \sum_{j \in \mathcal{E}_t} w_{t,j} (B_j A_j u_t).
\end{equation}
This design ensures that the model dynamically retrieves the most relevant expertise for each token without requiring explicit task identifiers. This substantially mitigates catastrophic forgetting while allowing for positive transfer, as vision-centric tokens can route to vision experts and logic-centric tokens to reasoning experts.

\section{Experiments}
This section details our experimental setup, including model architectures, datasets, and training hyperparameters. We evaluate our routing framework on InternVL-2~\cite{chen2024internvl} (2B and 8B variants) and LLaVA-v1.5-7B~\cite{liu2023visual} to assess scalability and robustness across diverse model families.

\begin{table*}[t]
\centering
\caption{Comparison of continual learning methods on InternVL2-2B and 8B. \textbf{Bold} indicates best performance; \underline{underline} indicates second best. Performance is reported across four sequential specialization tasks and three general-purpose benchmarks.}
\label{tab:combined-internvl-results}
\setlength{\tabcolsep}{3.5pt}
\begin{small}
\begin{tabular}{ll|cccc|ccc}
\toprule
\textbf{Size} & \textbf{Method} & \textbf{COCO} & \textbf{SNLI} & \textbf{HM} & \textbf{MMIMDb} & \textbf{MMB} & \textbf{ChartQA} & \textbf{DocVQA} \\
\midrule
\multirow{6}{*}{\textbf{2B}} 
& Zero-shot & 0.795 & 52.9 & 60.2 & 46.6 & \textbf{79.1} & \textbf{59.5} & \textbf{84.7} \\
& Seq-FT & 1.181 & 61.0 & 49.7 & 62.3 & 67.1 & 41.2 & 70.0 \\
& Arith. Merge & 1.171 & 73.3 & 64.5 & \textbf{64.9} & 78.6 & 59.2 & 84.1 \\
& MR-LoRA & \underline{1.310} & 80.0 & 69.1 & 61.4 & - & - & -  \\
\rowcolor[gray]{0.9}
& MTL & \textbf{1.340} & \textbf{83.6} & \textbf{73.9} & 63.0 & 76.5 & 55.7 & 77.9 \\
\rowcolor[gray]{0.9}
& \textbf{Routing (Ours)} & {1.305} & \underline{80.3} & \underline{71.2} & \underline{63.2} & \underline{78.8} & \underline{59.2} & \underline{84.3} \\
\midrule
\multirow{6}{*}{\textbf{8B}} 
& Zero-shot & 0.889 & 70.2 & 65.4 & 54.5 & 85.6 & \underline{72.6} & \textbf{89.7} \\
& Seq-FT & 1.258 & 65.4 & 52.1 & 64.8 & 71.2 & 48.5 & 74.3 \\
& Arith. Merge & 1.272 & 71.3 & 68.7 & 66.9 & 86.0 & 70.6 & 87.2 \\
& MR-LoRA & \underline{1.351} & 81.0 & 73.4 & 66.7 & - & - & - \\
\rowcolor[gray]{0.9}
& MTL & \textbf{1.374} & \textbf{84.7} & \textbf{78.3} & \underline{65.4} & \underline{85.7} & 68.2 & 86.5 \\
\rowcolor[gray]{0.9}
& \textbf{Routing (Ours)} & {1.350} & \underline{83.3} & \underline{76.8} & \textbf{67.4} & \textbf{86.3} & \textbf{73.1} & \underline{89.3} \\
\bottomrule
\end{tabular}
\end{small}
\end{table*}

\subsection{Datasets and Evaluation Metrics} 
To evaluate the stability-plasticity trade-off, we categorize our evaluation into \textit{Specialization Tasks} and \textit{General Benchmarks}. 

\paragraph{Specialization Tasks.} We employ a diverse set of multimodal (image+text) tasks covering: (1) \textbf{COCO-Caption} (COCO)~\cite{lin2014microsoft} for image captioning (CIDEr); (2) \textbf{SNLI-VE}~\cite{xie2019visual} for visual entailment (Accuracy); (3) \textbf{Hateful Memes} (HM)~\cite{kiela2020hateful} for toxicity detection (Accuracy); and (4) \textbf{MMIMDb}~\cite{arevalo2017gated} for movie genre classification (Precision).

\paragraph{General Benchmarks.} We monitor retention of foundational knowledge using \textbf{MMBench (MMB)}~\cite{liu2024mmbench} for multi-dimensional reasoning, \textbf{ChartQA}~\cite{masry2022chartqa} for visual data interpretation, and \textbf{DocVQA}~\cite{mathew2021docvqa} for document understanding.

\paragraph{MLLM-CL benchmark.}\cite{zhao2025mllmclcontinuallearningmultimodal} This benchmark evaluates the model’s ability to sequentially acquire knowledge across highly specialized and disparate domains, including Remote Sensing (RS), Medical imaging (Med), Art and Design (AD), Science (Sci), and Finance (Fin).

\paragraph{Cross-Modal Transfer.} To investigate the cross-modal transfer capabilities of the proposed method, we train two distinct experts: a text-only math expert (Orca-Math~\cite{mitra2024orca}) and a multimodal math expert (Multi-math~\cite{peng2024multimath}). We then evaluate zero-shot performance on the MGSM framework~\cite{shi2022language} in both English and Chinese.

\subsection{Training Hyperparameters}
\paragraph{Baseline Configurations.} For Sequential Finetuning (Seq-FT), we use a learning rate of $1\times10^{-5}$ for three epochs on each task. For Multi-Task Learning (MTL), all tasks are co-trained for three epochs. All experiments are conducted on 32 A10 GPUs with a global batch size of 512. Arithmetic Merging (Arith. Merge)~\cite{ilharco2022editing} computes a weighted average finetuned weights to create a single model.

\paragraph{Modular Routing Training.} We follow the two-stage training strategy as discussed in Section~\ref{sec:proposed_method}:
\begin{enumerate}
    \item \textbf{Expert Specialization:} Task-specific LoRA modules are trained ($lr=4\times10^{-5}$, cosine decay) within the LLM backbone.
    \item \textbf{Gate Optimization:} We freeze the LoRA weights and train the routing vector $v$ for 100 steps on the task data with batch size of 32 to map the expert to its specific activation manifold.
\end{enumerate}

\paragraph{Task Sequence.} We evaluate continual learning using the sequence: $\text{COCO} \rightarrow \text{SNLI} \rightarrow \text{HM} \rightarrow \text{MMIMDb}$. During inference, we employ top-$k$ routing with $k=1$, prioritizing the most relevant expert for each token to maintain computational efficiency. We follow the suggested tasks sequence for MLLM-CL benchmark~\cite{zhao2025mllmclcontinuallearningmultimodal}.

\section{Results and Analysis}
\label{sec:results_and_analysis}

\subsection{Performance Comparison and Stability}
As demonstrated in Table~\ref{tab:combined-internvl-results}, the proposed routing method (\textbf{Routing}) consistently achieves the optimal balance between task-specific specialization and the preservation of foundational capabilities.

\paragraph{Routing vs. Multi-Task Learning.}
Notably, our routing approach achieves performance comparably with the MTL model. On MMIMDb, routing surpasses MTL in both the 2B (63.2 vs. 63.0) and 8B (67.4 vs. 65.4) variants. This suggests that routing effectively mitigates \textit{negative task interference}~\cite{wang2020negative}. While MTL forces a shared weight space to accommodate contrasting data distributions, our approach isolates task-specific features, preventing destructive gradient interference.

Furthermore, routing demonstrates superior stability. MTL exhibits performance degradation on MMBench and DocVQA compared to the zero-shot baseline, indicating a forgetting of foundational knowledge. In contrast, routing preserves these scores near zero-shot levels.

\paragraph{Routing vs. Arith. Merge.}
Our experiments reveal a persistent performance gap between merging and routing on specialized tasks. While arithmetic merging effectively preserves general-purpose capabilities, it fails to fully recover the specialized performance of individual LoRA modules. For example, in the 8B setting, routing outperforms merging by $12.0\%$ on the SNLI task (83.3 vs. 71.3).

\paragraph{Routing vs. Seq-FT.}
Sequential FT exhibits the most severe catastrophic forgetting, with performance on general benchmarks like MMBench dropping sharply (from 79.1 to 67.1 in the 2B model). While merging and MTL offer varying degrees of stability, they either sacrifice peak task performance or require simultaneous access to all training data. Routing achieves a ``best-of-both-worlds'' outcome: it doesn't require access to past training dataset while matching or exceeding the stability of global multi-task training.

\begin{table}[t]
\centering
\setlength{\tabcolsep}{2pt} 
\caption{Comparison of routing vs. LoRA-specialized models. Routing achieves competitive or superior performance on all tasks using a single model. InternVL2-8B is used.}
\begin{scriptsize}
\begin{tabular}{l|cccc|ccc}
\toprule
\multirow{2}{*}{\textbf{Model}} 
& \multicolumn{4}{c|}{\textbf{Specialized Tasks}} 
& \multicolumn{3}{c}{\textbf{Benchmark Tasks}} \\
& COCO & SNLI & HM & MMIMDb & MMB & ChartQA & DocVQA \\
\midrule
Zero-Shot & 0.889 & 70.2 & 65.4 & 54.2 & \underline{85.6} & \underline{72.7} & \textbf{89.7} \\
\midrule
LoRA-COCO & \underline{1.348} & 71.8 & 63.2 & 47.8 & \textbf{86.3} & 71.5 & 89.4 \\
LoRA-SNLI & 1.084 & \textbf{84.2} & 63.0 & 53.2 & \underline{85.6} & 72.4 & \underline{89.6} \\
LoRA-HM & 1.004 & 74.0 & \underline{76.5} & 52.6 & 85.7 & 71.9 & 84.0 \\
LoRA-MMIMDb & 0.991 & 69.3 & 64.5 & \underline{65.8} & 85.8 & 72.1 & 89.4 \\
\rowcolor[gray]{0.9}
\textbf{Routing (Ours)} & \textbf{1.350} & \underline{83.3} & \textbf{76.8} & \textbf{67.4} & \textbf{86.3} & \textbf{73.1} & 89.4 \\
\bottomrule
\end{tabular}
\end{scriptsize}
\label{tab:benchmark_finetuning}
\end{table}

\paragraph{Comparison with MR-LoRA.} Our framework offers several key advantages over MR-LoRA~\cite{zhao2025mllmclcontinuallearningmultimodal}. Although MR-LoRA achieves strong task specialization, it requires a dual-inference pass, which increases computational overhead. In contrast, our routing mechanism is context-aware and completes inference in a single pass. Moreover, our approach benefits from \textit{cross-task synergy}: by operating at the token level, the router can dynamically leverage specialized knowledge from previously learned tasks to assist in new domains. This capability enables routing to outperform MR-LoRA.

\subsection{Specialization vs. Generalization}
A key question in modular architectures is whether a single routed model can match the performance of individual, fully specialized models. Table~\ref{tab:benchmark_finetuning} compares our framework against single-task LoRA experts.

The results indicate that routing does not merely match the performance of its experts; it often outperforms the individual LoRA variants from which it is composed. For example, the routed model achieves a score of 76.8 on Hateful Memes, surpassing the 76.5 achieved by the standalone LoRA-HM model. This suggests that the routing mechanism can perform "cross-expert utilization," where tokens within a single input benefit from the specialized features of multiple expert modules.

\begin{table}[!t]
\centering
\caption{Performance comparison on the MLLM-CL benchmark using LLaVA. Comparison is restricted to purely incremental approaches. Results are rounded to one decimal place for clarity.}
\label{tab:mllm_cl_benchmark}
\resizebox{\columnwidth}{!}{
\begin{tabular}{l|ccccc}
\toprule
\textbf{Method} & \textbf{RS} & \textbf{Med} & \textbf{AD} & \textbf{Sci} & \textbf{Fin} \\
\midrule
Zero-shot & 32.3 & 28.3 & 15.6 & 35.6 & 62.6 \\
Oracle & 81.1 & 65.8 & 54.2 & 56.9 & 91.1 \\
\midrule
LoRA-FT \cite{hu2022lora} & 69.7 & 41.6 & 25.4 & 40.9 & 87.5 \\
O-LoRA \cite{wang2023orthogonal} & 74.6 & 44.4 & 30.0 & 41.5 & 87.2 \\
MoELoRA \cite{chen2024internvl} & 77.5 & 41.9 & 27.6 & 40.1 & 86.8 \\
CL-MoE \cite{huai2025cl} & 71.3 & 46.8 & 26.3 & 41.2 & 88.7 \\
HiDe \cite{guo2025hide} & 74.3 & 49.0 & 33.2 & 38.5 & 81.6 \\
SEFE \cite{chen2025sefe} & 77.3 & 50.4 & 37.2 & 40.9 & 86.8 \\
DISCO \cite{guo2025federated} & 76.0 & 45.2 & 43.8 & 42.3 & 89.0 \\
ModalPrompt \cite{zengdual} & 53.6 & 45.7 & 40.8 & 41.8 & 87.8 \\
L2P \cite{wang2022learning} & 63.8 & 34.6 & 23.0 & 38.6 & \textbf{93.0} \\
\midrule
MR-LoRA \cite{zhao2025mllmclcontinuallearningmultimodal} & 80.9 & 65.3 & \textbf{54.1} & \textbf{56.7} & 91.1 \\
\rowcolor[gray]{0.9}
\textbf{Routing (Ours)} & \textbf{81.0} & \textbf{66.6} & 53.4 & 56.4 & 91.1 \\
\bottomrule
\end{tabular}
}
\end{table}

\subsection{Standardized Evaluation on MLLM-CL Benchmark}
To further validate our framework, we evaluate it against the MLLM-CL benchmark, which focuses on various domains such as Remote Sensing and Medical imaging. As summarized in Table~\ref{tab:mllm_cl_benchmark}, our routing mechanism effectively bridges the gap between zero-shot baselines and oracle.

The results highlight two primary strengths of our approach: First, by isolating domain-specific features into dedicated expert subspaces, our method achieves scores of 81.0 in Remote Sensing and 66.6 in Medical imaging, matching or slightly exceeding oracle performance. This indicates that modular specialization can be more effective than joint training, which often suffers from destructive gradient interference. Second, our approach maintains high efficiency by delivering competitive results—such as 91.1 in Finance and 56.4 in Science—using a single-pass inference. This offers a significant computational advantage over state-of-the-art baselines like MR-LoRA, which require expensive dual-inference passes to achieve similar performance.

Unlike traditional parameter-efficient continual learning methods, our framework does not require access to historical data or previous task gradients. This independence, combined with the ability to perform cross-expert utilization at inference time, confirms that token-level routing is a robust and scalable solution for multimodal continual learning.

\begin{table}[!t]
\centering
\caption{Performance of routing models with incremental inclusion of modules across datasets using InternVL2-2B. Zero-shot baseline included for reference.}
\label{tab:routing_models}
\resizebox{\columnwidth}{!}{
\begin{tabular}{l|c|c|c|c}
\toprule
\textbf{Routing Setup} & COCO & SNLI & HM & MMIMDb \\
\midrule
Zero-Shot & 0.795 & 52.9 & 60.2 & 46.6 \\
\midrule
SNLI, COCO & 1.306 & 80.0 & 61.9 & 48.3 \\
+ HM & \underline{1.309} & \underline{80.7} & \textbf{71.6} & 45.3 \\
+ MMIMDb & 1.305 & 80.3 & \underline{71.2} & \textbf{63.2} \\
+ ChartGemma & \textbf{1.311} & 80.3 & 70.8 & 62.8 \\
+ XNLI & 1.308 & \textbf{81.2} & \underline{71.2} & \underline{62.9} \\
\bottomrule
\end{tabular}}
\end{table}

\subsection{Ablation Studies}
In this section, we conduct a series of ablation experiments to evaluate the robustness, scalability, and computational efficiency of our routing-based framework.

\paragraph{Scalability and Expert Interference.} We first evaluate the stability of the routing mechanism as the expert pool grows. A critical challenge in modular architectures is ensuring that the addition of new task-specific experts does not induce performance degradation.

We begin with a base configuration routing between two experts: SNLI and COCO. As indicated in Table~\ref{tab:routing_models}, this setup yields substantial improvements on target tasks while maintaining near-baseline performance on unrelated tasks like Hateful Memes (HM) and MMIMDb.

The incremental addition of the HM module leads to a substantial gain in domain-specific performance (\textbf{60.2 → 71.6}), demonstrating the router's capacity to correctly activate the relevant expert. Similarly, including the MMIMDb module significantly improves performance in that domain (\textbf{46.6 → 63.2}) while maintaining the established performance levels of prior tasks.

To assess resilience against irrelevant expertise, we introduced a LoRA module trained on \textit{ChartGemma}~\cite{masry2025chartgemma}, a dataset unrelated to the core evaluation suite. As shown in Table~\ref{tab:routing_models}, the inclusion of this unrelated expert does not degrade performance across existing tasks, reinforcing the framework's scalability and its ability to ignore distractor modules.

Finally, we introduced an expert for XNLI~\cite{conneau2018xnli}, a text-only natural language inference task closely related to the multimodal SNLI. Interestingly, the addition of XNLI not only boosted performance on its own task but also improved scores on related image-text tasks (\textbf{80.3 → 81.2}). This suggests a \textit{positive transfer} effect, where the routing mechanism effectively leverages cross-modal similarities to enhance reasoning across related domains.

\begin{table}[t]
\centering
\caption{Computational efficiency comparison across methods. Training time measured in GPU hours on 32 A10 GPUs. Data requirements shown as relative factors, with the first task as baseline.}
\resizebox{\columnwidth}{!}{
\begin{tabular}{l|cc|cc}
\toprule
\textbf{Method} & \multicolumn{2}{c|}{\textbf{Training Time (hrs)}} & \multicolumn{2}{c}{\textbf{Data Requirements}} \\
 & First Task & All Tasks & First Task & All Tasks \\
\midrule
Seq-FT & 2.5 & 10.0 & 1× & 1× \\
Seq-FT+ER & 2.5 & 13.5 & 1× & 2.2× \\
MTL & 2.5 & 28.7 & 1× & 4× \\
\rowcolor[gray]{0.9}
\textbf{Routing (Ours)} & 2.5 & 10.6 & 1× & 1× \\
\bottomrule
\end{tabular}}
\label{tab:computational_efficiency}
\end{table}

\paragraph{Computational Efficiency Analysis.} We analyze the computational advantages of our routing-based approach compared to standard continual learning methods and existing routing baselines. Table~\ref{tab:computational_efficiency} summarizes training time and data requirements for InternVL2-2B across a four-task sequence.

As shown in Table~\ref{tab:computational_efficiency}, our framework is significantly more efficient than MTL and Experience Replay (ER). Beyond these standard baselines, our method offers two distinct advantages over recent modular approaches that utilize external router modules.

First, our method maintains data independence. Many existing routing frameworks require saving a subset of data~\cite{zhao2025mllmclcontinuallearningmultimodal} (e.g., 50 samples per task) from all past and current tasks to train a centralized router LoRA module. This creates a persistent dependency on historical data, violating the strict O(1) data scaling requirement of true continual learning. In contrast, our routing vectors are trained in isolation for each task, requiring zero access to prior data.

Second, our framework utilizes single-pass inference where based on input embeddings routing vector determine which LoRA module to be used. Methods such as MR-LoRA~\cite{zhao2025mllmclcontinuallearningmultimodal} methods often necessitate a "double inference" call: an initial pass to determine the appropriate expert via a router module, followed by a second pass to generate the response using the selected expert.

\subsection{Routing Enables Cross-modal Transfer}
Building upon the positive transfer observed between the text-only XNLI and multimodal SNLI tasks in Table~\ref{tab:routing_models}, we further investigate the effectiveness of routing in facilitating cross-modal synergy. We evaluate this capability on the multi-lingual math dataset MGSM~\cite{shi2022language}, specifically focusing on zero-shot reasoning performance.

As detailed in Table~\ref{tab:cross_modal_table}, our experimental setup utilizes two distinct LoRA modules: \textit{orca-math}, which is specialized for text-based mathematical reasoning, and \textit{multi-math}, which is trained on both image and text modalities. While the text-centric \textit{orca-math} expert naturally outperforms the multimodal variant on this specific task, the routing mechanism achieves a superior aggregate performance (0.82 vs. 0.78 for English).


\begin{table}[t]
\centering
\caption{Cross-modal experiments on the MGSM dataset. Routing leverages modality-specific experts (Text-only vs. Vision + Text) to achieve superior performance, demonstrating successful knowledge transfer.}
\label{tab:cross_modal_table}
\resizebox{\columnwidth}{!}{
\begin{tabular}{l|c|cc|c}
\toprule
\multirow{2}{*}{\textbf{Task}} & \multirow{2}{*}{\textbf{Zero-shot}} & \multicolumn{2}{c|}{\textbf{Expert Models}} & \multirow{2}{*}{\textbf{Routing}} \\
\cmidrule(lr){3-4}
& & \textit{orca-math} (T) & \textit{multi-math} (V+T) & \\
\midrule
English & 0.57 & \underline{0.78} & 0.62 & \textbf{0.82} \\
Chinese & 0.57 & \underline{0.62} & 0.58 & \textbf{0.64} \\
\bottomrule
\end{tabular}
}
\end{table}

Notably, the routed model often surpasses the performance of individual specialized experts, demonstrating that our framework does not merely select the most relevant module but dynamically synthesizes complementary information across modalities. By integrating the specialized reasoning of text-only experts with the alignment capabilities of multimodal experts, the routing mechanism effectively expands the model's performance.
\section{Discussion}
While our routing approach shares architectural similarities with Mixture of Experts (MoE) systems like MoE-LLaVA \cite{lin2024moe}, it fundamentally differs in both purpose and implementation. Traditional MoE approaches improve model capacity through joint training of specialized experts within a unified architecture, requiring simultaneous access to all training data and joint optimization of experts and gating networks.

Our method diverges from this paradigm in three key aspects. First, our experts (LoRA modules) are trained independently and sequentially, with each expert specializing on a single task without access to other tasks' data. Second, we explicitly decouple expert learning from routing by first training task-specific LoRA modules, then learning lightweight routing vectors while keeping LoRA parameters frozen, unlike traditional MoEs where expert structure and routing are jointly optimized. Third, our approach requires only new task data, eliminating the need to retain or synthesize previous task data.

\section{Conclusion}
In this work, we address the challenge of catastrophic forgetting in MLLMs by introducing a token-level routing framework that seamlessly integrates new task-specific experts while strictly preserving foundational knowledge. Our approach distinguishes itself from traditional multi-task learning and experience replay by maintaining fixed resource requirements per task and eliminating the dependency on historical data, thus overcoming the linear computational growth that typically limits large-scale continual learning. Through extensive evaluations, we demonstrate that routing achieves performance parity with the theoretical multi-task upper bound on specialized tasks while simultaneously protecting general-purpose reasoning capabilities. Notably, we uncover that this modular architecture facilitates cross-modal transfer, allowing unimodal expertise to bolster multimodal reasoning, and exhibits a positive scaling trend where larger models demonstrate superior resistance to forgetting. Unlike existing routing baselines that require multi-pass inference and data replay, our single-pass mechanism provides an efficient and scalable solution for real-world deployment. Future research will explore the extension of this modularity to support cross-lingual transfer in massively multilingual multimodal settings.
\section{Limitations}
\label{sec: Limitation}
While our routing-based framework offers a scalable solution for multimodal continual learning, some limitations merit further investigation. First, the current implementation necessitates the simultaneous loading of multiple LoRA adapters during inference to enable token-level dynamic composition. Although each adapter is parameter-efficient , the cumulative memory footprint scales linearly with the number of tasks, which may present deployment challenges on hardware with restricted VRAM. Future work could explore hot-swapping mechanisms to dynamically load and unload adapters from system RAM to GPU memory based on high-level task identifiers or predicted routing requirements, thereby maintaining a near-constant inference memory overhead. Second, our training strategy prioritizes stability by isolating experts, which inadvertently minimizes sequential transfer during the training phase. Because each LoRA module is trained independently on its respective task without access to prior adapters, the model cannot inherently leverage "forward transfer" to accelerate the acquisition of subsequent, related skills during the optimization process. While we observe positive transfer at inference time through token-level routing , the training process remains strictly decoupled. Further research is needed to determine if initializing new experts via knowledge distillation from existing ones could facilitate collaborative learning without reintroducing catastrophic forgetting.
\bibliography{references}

\newpage
\appendix
\onecolumn
\section{Dataset description and sizes}
\label{Appendix: Dataset Sizes}
For our fine-tuning experiments, we have selected a diverse set of datasets that target distinct tasks and test various facets of multimodal learning. These datasets span a range of applications, including image captioning, visual entailment, hate speech detection, and multimodal classification, providing a comprehensive evaluation framework for our models. Below are the datasets used in our experiments:

\textbf{COCO-Caption} dataset consists of over 330,000 images, each annotated with five human-generated captions, originally designed for image captioning tasks. We use an adapted split, which includes 566,747 image-text pairs for training and 5,000 image-text pairs for testing. This provides a rich multimodal resource for exploring the relationships between visual content and natural language descriptions. The performance is reported using the CIDEr~\cite{vedantam2015cider} metric, which evaluates the quality of generated captions based on their similarity to reference captions.

\textbf{SNLI-VE} is derived from the Stanford Natural Language Inference (SNLI)~\cite{bowman2015large} corpus, SNLI-VE (Visual Entailment) extends this dataset by including visual contexts. It consists of 529,527 image-text pairs for training, with a test set of 2,000 pairs. These pairs are categorized into three labels: (a) entailment, (b) neutral, and (c) contradiction, allowing models to evaluate their ability to reason about the relationships between visual and textual information. We use accuracy as the evaluation metric for this task.

\textbf{Hateful Memes} is multimodal dataset is specifically designed to evaluate models' ability to detect hateful content. It contains 8,500 training examples and 1,000 test examples, with each example combining text and images. The task requires models to interpret both modalities to accurately identify hate speech. The dataset includes two labels: (a) not-hateful and (b) hateful. Accuracy is used as the performance metric for this dataset.

\textbf{MMIMDb} is aimed at evaluating multimodal movie genre classification. It contains pairs of movie plot texts and posters, with 15,552 training examples and 2,592 testing examples. The dataset is similar to a multilabel classification task, with 25 possible genre labels. We evaluate the models using the precision metric, which measures the proportion of true positive predictions out of all positive predictions.

These datasets represent a broad spectrum of multimodal tasks, enabling us to assess our models' ability to handle diverse challenges that involve integrating both visual and textual information. We utilize a variety of performance metrics—such as CIDEr, accuracy, and precision—to comprehensively evaluate the models' capabilities across different tasks.

\subsection{Benchmarks}
In addition to evaluating metrics on fine-tuning tasks, we conduct an in-depth analysis of the fine-tuned model using well-established benchmarks to assess the extent of forgetting introduced during model training. For this analysis, we utilize the MMBench~\cite{liu2024mmbench}, ChartQA~\cite{masry2022chartqa}, and DocVQA~\cite{mathew2021docvqa} benchmarks, each targeting distinct evaluation objectives.

\textbf{MMBench} is designed to assess a model's performance on a set of fine-grained abilities, encompassing 20 ability dimensions under perception and reasoning. It uses multiple-choice questions that challenge the model to demonstrate both fine-grained understanding and reasoning skills.

\textbf{ChartQA} evaluates the model's ability to perform complex reasoning over data visualizations such as charts and graphs. The questions in this evaluation often require logical and arithmetic reasoning, as well as an understanding of the visual features of the chart, which tests a model’s ability to interpret and reason about structured data.

\textbf{DocVQA} Focused on visual question answering (VQA) tasks involving document images, this dataset requires models to comprehend and reason over the structure and content of documents. It assesses the model's ability to interpret visual and textual information in the context of documents, a challenge unique to VQA in document-specific scenarios.

These benchmark datasets are crucial for evaluating the robustness and generalization capabilities of our models, particularly in scenarios that involve more complex reasoning or domain-specific tasks, such as visual question answering in documents and data visualization comprehension. Additionally, they provide a means of assessing the potential for CF in the fine-tuned models, ensuring their continued effectiveness across various domains.

\section{Routing patterns}

\begin{figure*}[ht]
\vskip 0.2in
\begin{center}
    \begin{minipage}[b]{0.32\textwidth}
        \centering
        \includegraphics[width=\linewidth]{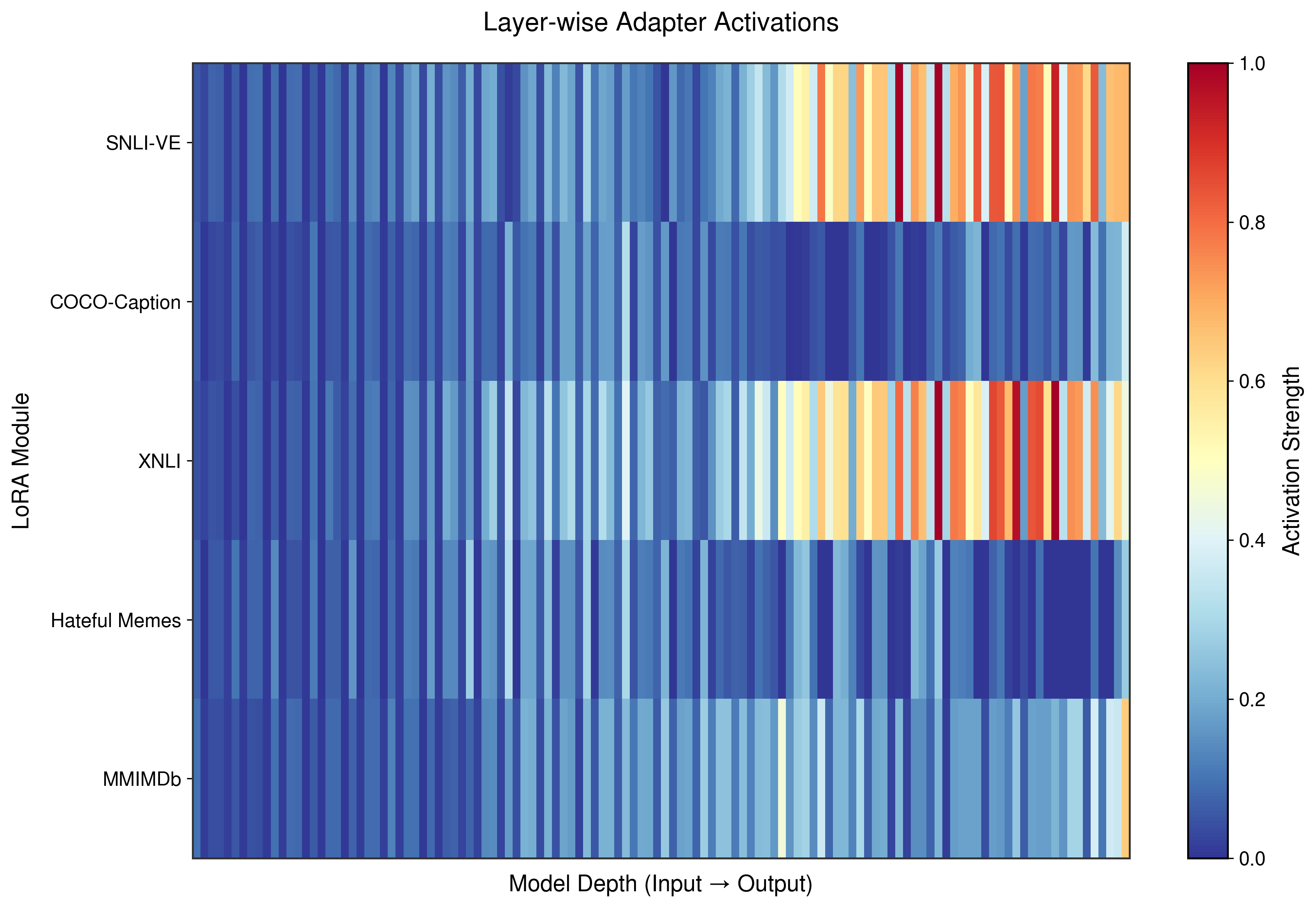}
        \label{fig:snli}
    \end{minipage}
    \hfill
    \begin{minipage}[b]{0.32\textwidth}
        \centering
        \includegraphics[width=\linewidth]{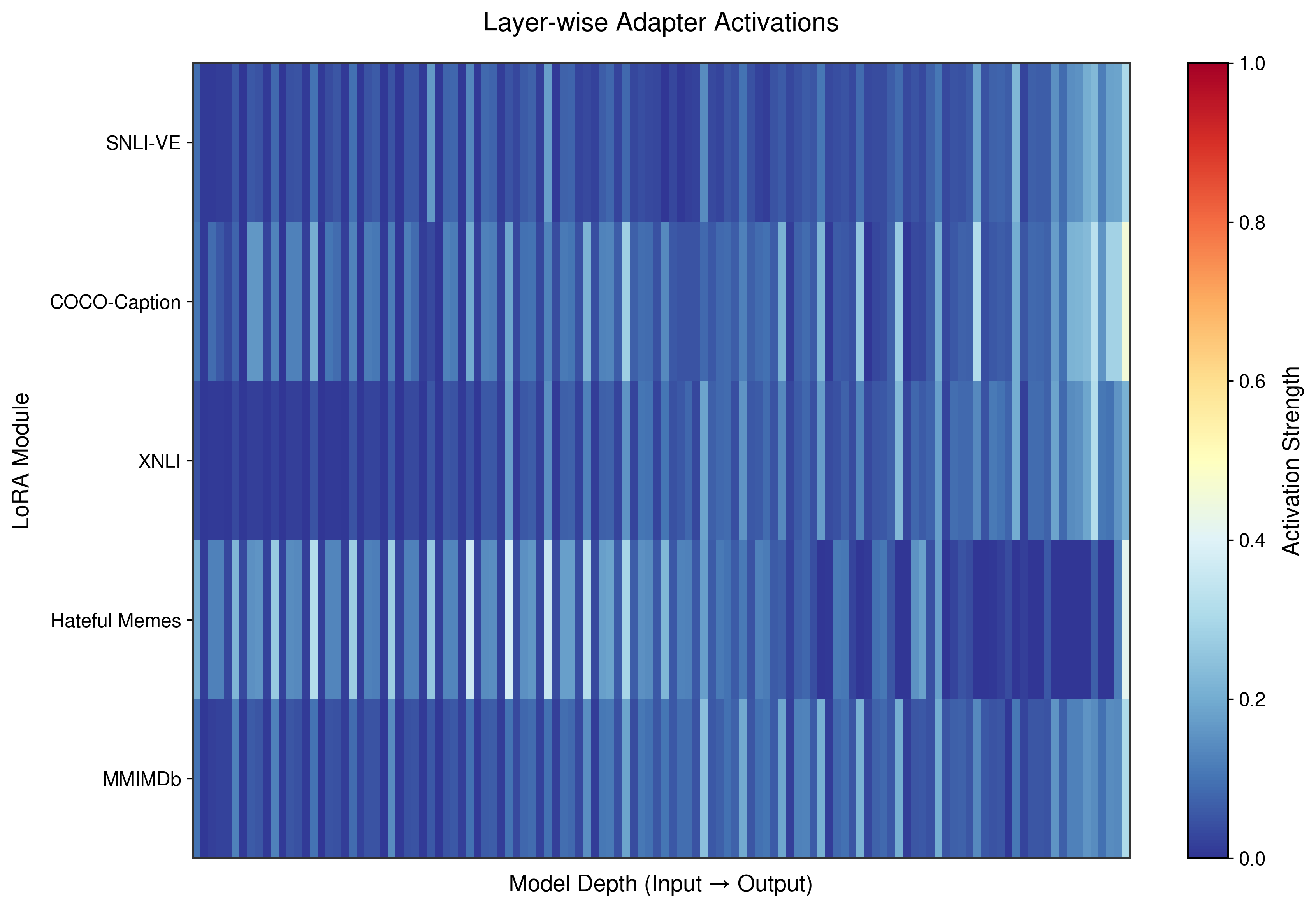}
        \label{fig:mmbench}
    \end{minipage}
    \hfill
    \begin{minipage}[b]{0.32\textwidth}
        \centering
        \includegraphics[width=\linewidth]{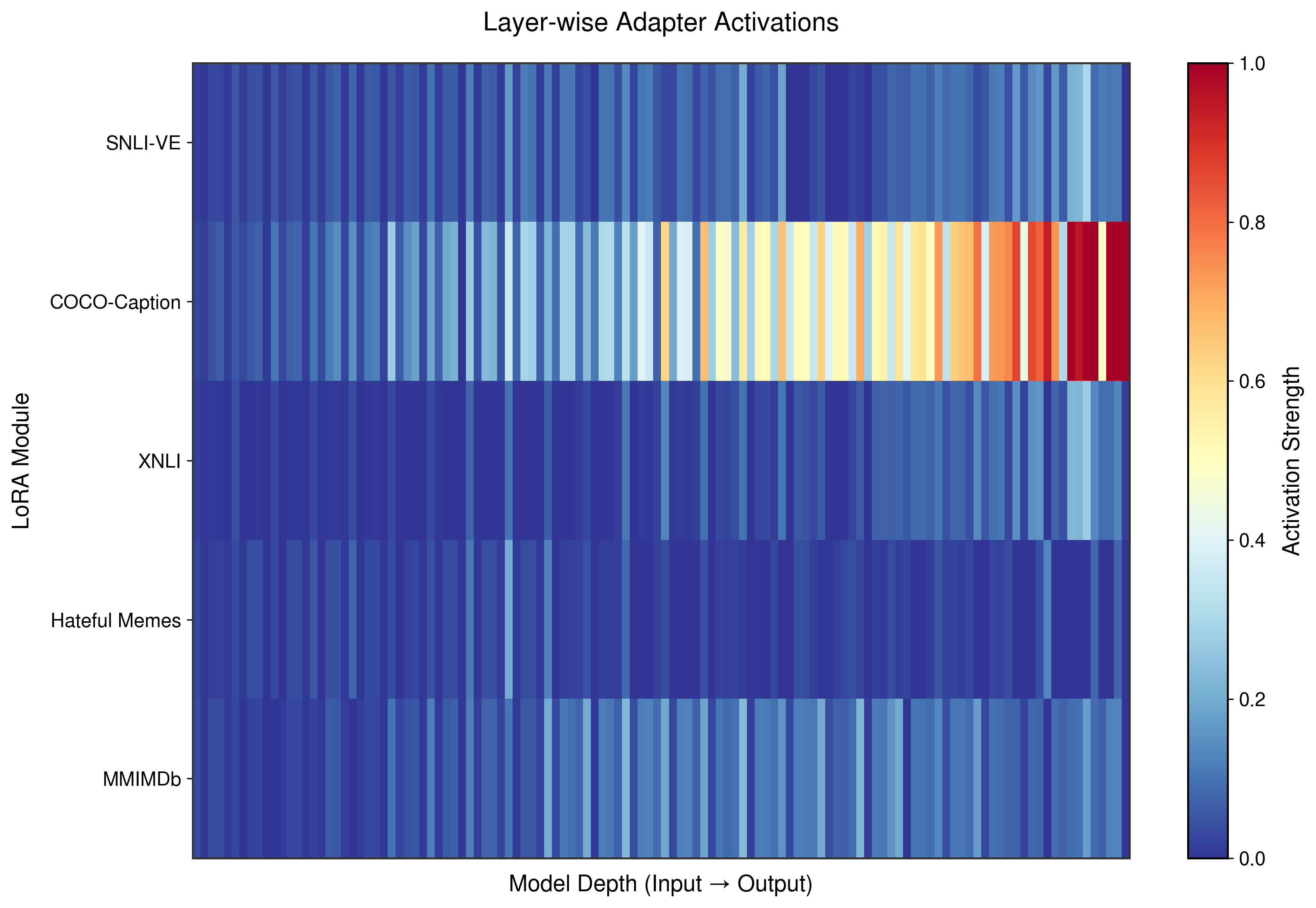}
        \label{fig:another}
    \end{minipage}
    \caption{Routing patterns for SNLI (left), MMBENCH (middle), and COCO (right). The figure demonstrates that the routing model correctly activates task-specific modules: SNLI and XNLI for SNLI, only COCO for COCO, and no module for MMBENCH, where the base model handles the task. These confirm the routing mechanism's effectiveness in selecting the appropriate module based on the task.}
    \label{fig:global_figure}
\end{center}
\vskip -0.1in
\end{figure*}
In Figure~\ref{fig:global_figure} (left), we examine the routing pattern for the SNLI task. The results clearly indicate that the SNLI module is correctly activated. However, we also observe the activation of the XNLI module, which highlights the ability of the routing vector to understand task similarity. This is intuitive, as XNLI is a text-text entailment task, while SNLI is a text-image entailment task. This pattern is consistent with the findings discussed in section on routing vs specialized model, where we observed that adding the XNLI module enhanced the performance of the SNLI task, likely due to shared underlying features between the two tasks. In Figure~\ref{fig:global_figure} (right), we present the routing pattern for the COCO task. Here, we observe that only the COCO module is activated, demonstrating that the routing vectors are correctly trained. This ensures that requests are routed automatically and accurately to the appropriate modules, validating the robustness of the routing.

In Figure~\ref{fig:global_figure} (middle), we present the routing pattern for the benchmark task MMBENCH. As this is a general-purpose task, we observe that no LoRA module is activated, and the model relies on the base model's knowledge to solve the task. This demonstrates the model’s ability to bypass LoRA modules when the base model has sufficient expertise to address a specific task, showcasing the efficiency and adaptability of the routing mechanism.

\section{Routing in cross-lingual setting}
\label{subsec: routing cross lingual transfer}
In this section, we investigate whether routing can enable cross-lingual transfer—specifically, whether a routing model can solve a task in one language by combining a LoRA module trained on that language with another LoRA module trained on a specialized task in English. To evaluate this, we focus on the MGSM dataset~\cite{shi2022language}, a multilingual math-based reasoning benchmark. We construct two expert models: a Math Expert, trained on the OrcaMath dataset~\cite{mitra2024orca}, and a Chinese Language Expert, trained on the xP3mt dataset~\cite{muennighoff2023crosslingual}.

Table~\ref{tab:multilingual_zh} presents results for InternVL2-2B and InternVL2-8B models. The 2B model benefits from routing, achieving the highest score when both the Math and Chinese experts are combined, demonstrating effective cross-lingual transfer. However, the 8B model struggles with routing, showing a significant performance drop when both experts are used together.

While the 8B model demonstrates strong zero-shot performance initially, the Chinese expert negatively affects the results, indicating potential interference with the model's pre-trained multilingual capabilities. The Math expert improves performance in isolation, but the routing mechanism fails to effectively combine both experts, causing a significant decline in performance. This suggests that the larger-scale language expert is underperforming, which in turn hampers the routing process. A more optimized data mixture for the language model may be necessary to address this issue.

\begin{figure*}[ht]
\vskip 0.2in
\begin{center}
    \begin{minipage}[b]{0.49\textwidth}
        \centering
        \includegraphics[width=\linewidth]{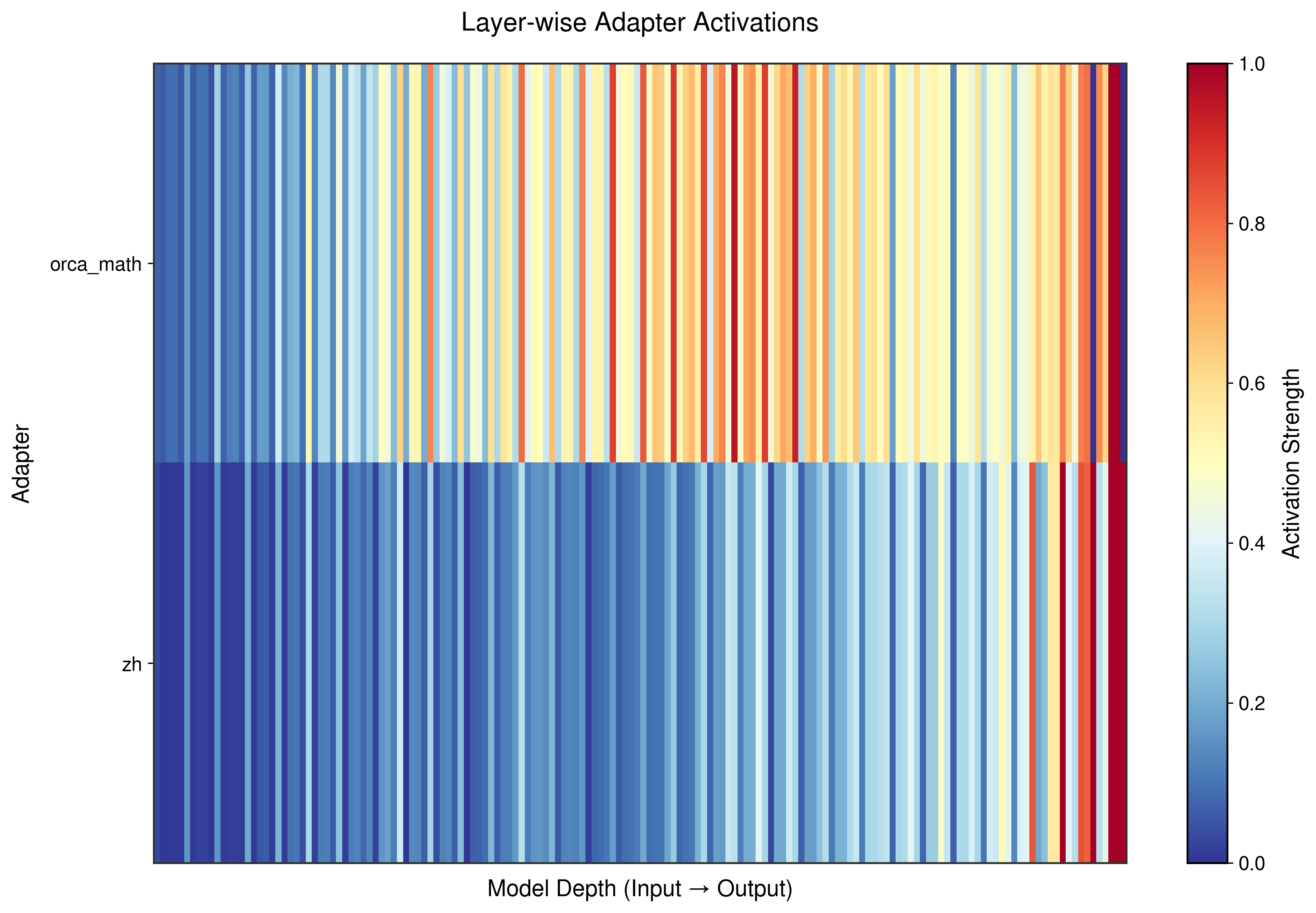}
        \label{fig:chinese}
    \end{minipage}
    \hfill
    \begin{minipage}[b]{0.49\textwidth}
        \centering
        \includegraphics[width=\linewidth]{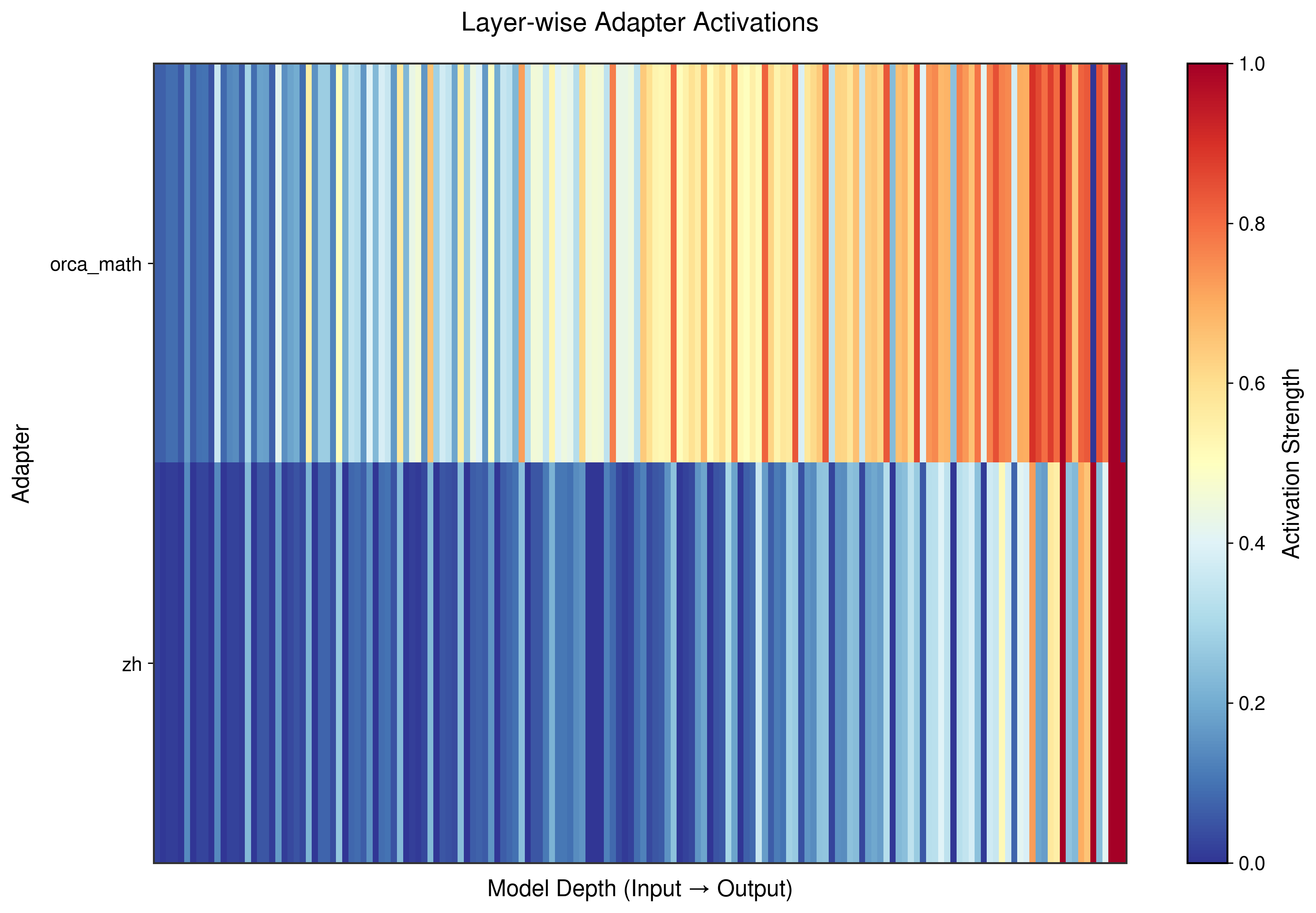}
        \label{fig:another}
    \end{minipage}
    \caption{Routing patterns for the MGSM dataset in multilingual transfer. Notably, the model leverages the Chinese expert in the early and later stages, while the math expert is primarily used for computation.}
    \label{fig:global_figure_zh}
\end{center}
\vskip -0.1in
\end{figure*}

Figure \ref{fig:global_figure_zh} reveals an interesting pattern in cross-lingual transfer. As expected, the language model is more frequently utilized in the initial and later layers, while the math expert plays a dominant role in performing computations. We believe this aligns with the expected behavior, where the initial layers transform embeddings from language to math, and the later layers map from the math space back to language.

\begin{table}[ht]
\caption{Comparison of routing models for cross-lingual transfer on the MGSM 8-shot dataset, evaluating different scales of the InternVL2 model on the Chinese subset.}
\vskip 0.2in
\begin{center}
\begin{small}
\renewcommand{\arraystretch}{1.2} 
\begin{tabular}{l|c|c|c|c}
\toprule
\multirow{2}{*}{\textbf{Model Scale
}} & \multirow{2}{*}{{zero-shot}} & \multicolumn{2}{c|}{\textbf{Expert Models}} & \multirow{2}{*}{{Routing}} \\
& & {Chinese} & {Math} & \\
\midrule
InternVL2-2B & 0.15 & 0.16 & \underline{0.20} & \textbf{0.22} \\
InternVL2-8B & 0.57 & \underline{0.53} & \textbf{0.62} & 0.47 \\
\bottomrule
\end{tabular}
\end{small}
\label{tab:multilingual_zh}
\end{center}
\vskip -0.1in
\end{table}

\section{Training details}
\label{appendix: training details}
For the full finetuning, we use a learning rate of $1\times10^{-5}$ and train the models for a single epoch, unless stated otherwise. We employ total of 32 A10 GPUs for training these models. The batch size for training is set to 512, and we freeze the ViT and projection layers, training only the language model’s weights. 

The models are trained using a cosine learning rate scheduler to adjust the learning rate over the course of training. We leverage the ZeRO-3 configuration to efficiently offload gradients, optimizer states, and parameters. The models are trained using a cosine learning rate scheduler to adjust the learning rate over the course of training. For specializing LoRA module, we employ a learning rate of $4\times10^-5$. LoRA modules are added exclusively to the LLM, ensuring that only the LoRA weights of the LLM are updated during training. To train the gating module we freeze all the parameters and only train the gate vector $v$ for 100 steps on the specialized task.

For sequential learning, we follow the task progression: COCO-Caption → SNLI-VE → Hateful Memes → MMIMDb. This sequence allows us to analyze the impact of each approach across diverse datasets and tasks.

\end{document}